\documentclass[journal]{IEEEtran}
\usepackage{graphicx}
\usepackage{subfig}
\usepackage{subfloat}
\usepackage{amsmath}
\usepackage{booktabs}
\usepackage{autobreak}
\allowdisplaybreaks
\ifCLASSINFOpdf
\else
\fi
\begin{document}
\title{PINNup: Robust neural network wavefield solutions using frequency upscaling and neuron splitting}

\author{{Xinquan Huang, and Tariq Alkhalifah}
\thanks{{\it(Corresponding author: Xinquan Huang)}\\
X. Huang and T. Alkhalifah are with Physical Sciences and Engineering Division, King Abdullah University of Science and Technology (email: xinquan.huang@kaust.edu.sa, tariq.alkhalifah@kaust.edu.sa).}}

\markboth{}%
{Huang \MakeLowercase{\textit{et al.}}.}

\IEEEtitleabstractindextext{%
\begin{abstract}
Solving for the frequency-domain scattered wavefield via physics-informed neural network (PINN) has great potential in seismic modeling and inversion. However, when dealing with high-frequency wavefields, its accuracy and training cost limits its applications. Thus, we propose a novel implementation of PINN using frequency upscaling and neuron splitting, which allows the neural network model to grow in size as we increase the frequency while leveraging the information from the pre-trained model for lower-frequency wavefields, resulting in fast convergence to high-accuracy solutions. Numerical results show that, compared to the commonly used PINN with random initialization, the proposed PINN exhibits notable superiority in terms of convergence and accuracy and can achieve neuron based high-frequency wavefield solutions with a two-hidden-layer model.
\end{abstract}

}
\maketitle

\IEEEdisplaynontitleabstractindextext
\IEEEpeerreviewmaketitle

\section{Introduction}
\IEEEPARstart{F}{requency}-domain seismic modeling, based on the Helmholtz wave equation, provides compact representations of subsurface wavefields \cite{Marfurt1984}. Those representations are highly desirable in applications like waveform inversion \cite{GerhardPratt1998}. However, the computational cost of numerical attaining such a solution increases exponentially with frequency. The cost becomes more of a burden when we have to compute multi high frequency wavefields, as each frequency often requires its own matrix inversion. Besides, the complexity of the wave equation in complex media like anisotropic ones also limits its use in applications like migration or full waveform inversion \cite{Wu2018}. 

A recently developed physics-informed neural network (PINN) framework proved to be a great tool for an efficient surrogate modeling for frequency-domain wavefields \cite{Alkhalifah2020a,Sitzmann2020,Song2021}. It can be used to solve for the wavefield in a self-supervised manner using physical constraints. The PINN solutions is adaptable to any model shape including when we have irregular topography. However, its implementation for high-frequency wavefield representations, in which the wavefields tend to be more complex, poses a challenge to the PINN framework. The universal approximation theorem \cite{Hornik1989} states that using neural network (NN) we can theoretically represent any continuous function with a sufficiently wide and deep enough NN. The model architecture, training samples, and even the initialization of the NN will affect the convergence and accuracy of the wavefield solution \cite{Alkhalifah2020a}. Previous work on PINN using positional encoding (PE) \cite{huang2021pinnpe} can improve the accuracy and accelerate convergence to some extent, but for high-frequency and even multi high-frequency wavefields, it still requires training the NN for every frequency with random initializations, which is costly. Alkhalifah {\it et al.} \cite{Tariq2021pinnfre} proposed high-dimensional solutions based on PINN, which trains the NN for multi-frequency wavefields simultaneously and predicts both low- and high-frequency wavefield. Its accuracy is limited because the NN needed to capture the features of both low- and high-frequency wavefields is not easy to train. 

Fortunately, NN has a low-frequency bias property \cite{Rahaman2018}, which means in the training of NN, the network first learns the low frequency components of a function and then slowly adds the higher frequency information. Since wavefields for a fixed velocity model share properties over many frequencies, like speed of propagation (general shape), we can use this feature in the NN training. Specifically, we train a neural network to predict low-frequency wavefields, and use the trained model to initialize the training for higher-frequencies, which we will refer to as {\it frequency upscaling in PINN} (PINNup). Now we ask ourselves, what network size can accommodate all the frequencies of interest? To capture the features embedded in the often complex wavefields at high-frequency and at the same time benefit from pretraining, we will also need a large neural network for low-frequency wavefield representations, which is a waste. Thus, in our proposed PINNup, we train an NN with a shallow and narrow architecture for low-frequency wavefields, and increase the size of our network as we upscale to high frequency. There are many ways to increase the number of neurons in each layer, like adding new neurons estimated by gradient boosting while keeping the previous neurons fixed \cite{Schwenk2000,bengio2006convex,Bach2017}, or adding new neurons with random initialization \cite{Chen2016}. However, these methods require more time to converge and do not benefit from the pretrained model. So we use a neuron splitting strategy \cite{liu2019splitting}, which can leverage the existing model for faster convergence. 

To summarize, the contribution of our work includes:
\begin{itemize}
  \item A novel PINN framework using frequency upscaling and neuron splitting, we refer to it as PINNup.
  \item An empirical formula that relates the neuron splitting to the frequency upscaling that allows for high accuracy and fast convergence.
  \item The proposed method can offer a compressed representation of wavefields.
\end{itemize}
In the following sections, we first introduce the theory of our proposed method, then compare its performance with the conventional approach on a simple layered model extracted from the Marmousi, which demonstrates the ability of our proposed method to efficiently represent more complex wavefields accurately.
\section{Methodology}
In the following subsections, we will review the wave equation for a scattered wavefield, then illustrate the idea of frequency upscaling within the framework of PINN, and finally share the general idea behind neuron splitting and its implementation. 
\subsection{The acoustic wave equation in frequency domain}
\label{sec:helm}
One major characteristic of wavefields in the frequency domain is that for individual frequencies these wavefields share the same general shape when they correspond to the same velocity model and source location. In the spirit of frequency strategies in full-waveform inversion (FWI), which optimizes the objective function sequentially from low to high frequency, we propose to optimize the neural network wavefield solution the same way. The proposed strategy, in principal, can help any wave equation in the frequency domain, {\it e.g.} acoustic/elastic wave equation in anisotropic/isotropic media even with attenuation. To develop the concept, we focus, in this paper, on the frequency upscaling and neuron splitting for the 2-D frequency-domain acoustic wave equation (Helmholtz equation), which is given by:
\begin{equation}
  (\omega ^2\mathbf{m} + \nabla ^2)\mathbf{U}(\mathbf{x}) = \mathbf{s},
  \label{fg}
\end{equation}
where $\mathbf{m}$ is the squared slowness, $\omega$ is the angular frequency, $\mathbf{U}$ is the frequency-domain wavefield as a function of $\mathbf{x}=(x,z)$ due to a source $\mathbf{s}=(s_x,s_z)$ and $\nabla$ is the gradient operator. To mitigate the influence of the source singularity and decrease the need of training samples, we consider the scattered wavefield $\delta \mathbf{U}=\mathbf{U}-\mathbf{U}_0$ \cite{Alkhalifah2020} in this paper. Then, the wave equation for scattered wavefield $\delta \mathbf{U}$ can be formulated as:
\begin{equation}
  \omega ^2\mathbf{m}\delta \mathbf{U} + \nabla ^2\delta \mathbf{U} + \omega^2\delta\mathbf{m}\mathbf{U}_0 = 0,
  \label{sg}
\end{equation}
where $\mathbf{U}_0$ is the background wavefield, and $\delta \mathbf{m}=\mathbf{m}-\mathbf{m}_0$ is the squared slowness perturbation. In this equation, $\mathbf{U}_0$ can be calculated based on a constant background squared slowness $\mathbf{m}_0$ using an analytical formula \cite{Richards1980}:
\begin{equation}
  \mathbf{U}_0({x,z}) = \frac{i}{4}{\it H_0^{(2)}}(\omega\sqrt{m_0\{(x-s_x)^2+(z-s_z)^2\}}
  \label{anal}
\end{equation} 
where ${\it H_0^{(2)}}$ is the zero-order Hankel function of the second kind.
\subsection{Frequency upscaling of wavefield representation in PINN}
\label{sec:2.1}
Learning an effective representation $\Phi(\theta, \mathbf{x})$ for a frequency-domain wavefield using a physical constraint (known as physics informed neural network, PINN) as a surrogate modeling technique is attractive. Here, $\theta$ represents the NN model parameters and $\mathbf{x}$ includes the input coordinates for the wavefield and the source location (dimensions of the Green's function). Taking a fully-connected deep neural network with $L$ layers as an example, the latent representation learned by the $\ell$-th layer, $\mathbf{H}^{(\ell)}$, is given by:
\begin{equation}
  \label{pinn}
  \mathbf{H}^{(\ell)}=\phi\left(\mathbf{W}^{(\ell)} \mathbf{H}^{(\ell-1)}+\mathbf{b}^{(\ell)}\right), \ell=1,...,L-1,
\end{equation}
where $\mathbf{W}^{\ell}$ and $\mathbf{b}^\ell$ are the weight matrices and bias vectors of the $\ell$-th layer, respectively, and $\phi$ is an activation function. Suggested by \cite{Sitzmann2020,huang2021pinnpe}, to improve the convergence and accuracy, the activation function we are using here is a sine activation function. Moreover, our previous research suggests that using positional encoding can improve the representation of NNs for frequency-domain wavefields \cite{huang2021pinnpe}. Thus, we set $\mathbf{H}^{(0)}$ as the input coordinates $\mathbf{x}=x,z,s_x$ (in 2D, and sources on the surface) and its positional embedded vector. 

Then the mapping function $\Phi$ can be formulated from the last linear layer as
\begin{equation}
  \label{pinn-full}
  \Phi(\theta,\mathbf{x}) = \mathbf{W}^{(L)}\mathbf{H}^{(L-1)}+\mathbf{b}^{(L)},
\end{equation}
and $\theta$ is the set of $\{\mathbf{W}^{(0)},\mathbf{b}^{(0)};\mathbf{W}^{(1)},\mathbf{b}^{(1)};\dots;\mathbf{W}^{(L)},\mathbf{b}^{(L)}\}$. 

During the training process, we use the wave equation (as a loss measure) to optimize the neural network, which maps the the input coordinates to scattered wavefield satisfying equation \ref{sg}. Thus, the corresponding loss function in this self-supervised training is given by
\begin{equation}
  \mathcal{L} = \frac{1}{N} \sum_{i=1}^{N}\left|\omega^{2} m^{i} \Phi(\theta, \mathbf{x}^i)+\nabla^{2} \Phi(\theta, \mathbf{x}^i)+\omega^{2} \delta m^{i} U_{0}^i\right|_{2}^{2},
  \label{loss} 
\end{equation}
where $i$ and $N$ represent the training sample index and the number of training samples $\mathbf{x}$, respectively. 

\begin{figure*}[!htb]
  \centering
  \includegraphics[width=0.8\textwidth]{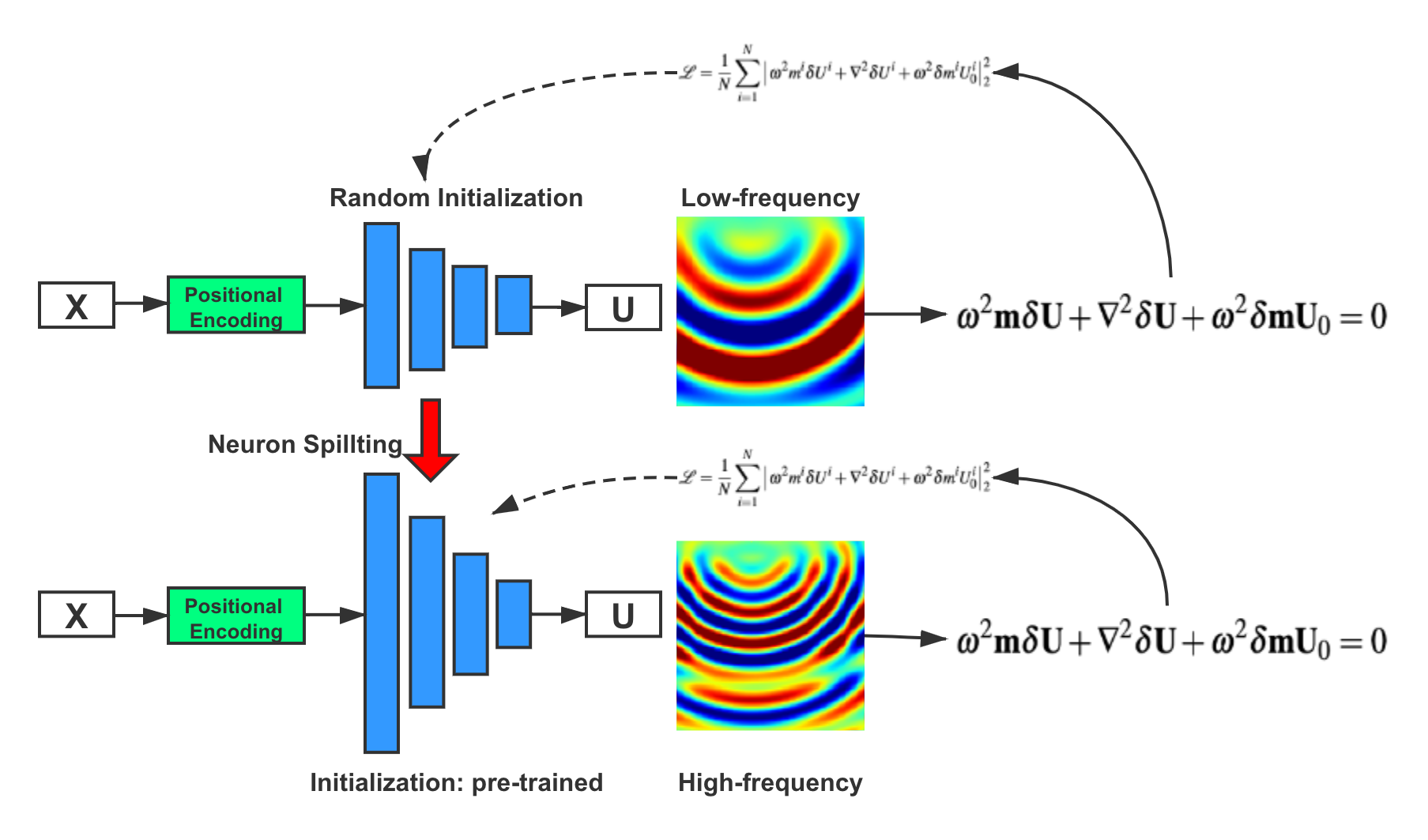}
  \caption{The framework of our proposed method, where $\mathbf{X}$ is the set of inputs $\mathbf{x}$, and $\mathbf{U}$ is the corresponding wavefield (the output). The basic NN architecture used here is PINN with sine activation and positional encoding \cite{huang2021pinnpe}.}
  \label{workflow}
\end{figure*}

Similar to FWI, our NN is first trained for low-frequency wavefields, and gradually optimized for high-frequency wavefields using the lower frequency NN parameters to initialize the model. The information contained in the NN for low-frequency wavefield is beneficial for higher-frequency wavefield training, leading to faster convergence and better prediction than training from scratch. From the prospective of deep learning, our source domain (low-frequency wavefield) and target domain (a higher-frequency wavefield) are inherently related through kinematic properties for a given velocity model and source location. However, higher frequency wavefields are dynamically more complex, and thus, will require neural network models (larger) that can represent the complex features. The workflow of the proposed method is illustrated in Figure~\ref{workflow}. After we train the neural network to predict low-frequency wavefields using a small network, we increase the size of the network through neural splitting and then use it to learn higher frequency representation. Thus, we can benefit from the NN training experience at low frequency to help us converge faster at high frequencies.
\subsection{Neuron splitting operation}
\label{sec:2.2}
As mentioned before, we want to leverage the information from trained models, while increasing the convergence speed of larger NNs needed for higher frequency representations. The concept of neuron splitting allows us to increase the network size without effecting its output \cite{liu2019splitting}. The splitting process of the neurons in a hidden layer involves duplicating the weights coming into the neuron to all of its off springs, while dividing the weights connecting the neuron with the next layer by the number of offsprings. In the case of splitting all the neurons, the weights and biases follow the following formulas: 
\begin{equation}
  \label{equ:inputl}
  \mathbf{W}^{(1)}_{split} = \begin{bmatrix} \mathbf{W}^{(1)} \dots \mathbf{W}^{(1)} \end{bmatrix}^T, \mathbf{b}^{(1)}_{split} = \begin{bmatrix} \mathbf{b}^{(1)}\dots\mathbf{b}^{(1)}\end{bmatrix}^T,
\end{equation}
\begin{equation}
  \label{equ:hidl}
  \mathbf{W}^{(\ell)}_{split} = \frac{1}{n}\begin{bmatrix} \mathbf{W}^{(\ell)} \dots \mathbf{W}^{(\ell)} \\
  \vdots  \\
  \mathbf{W}^{(\ell)} \dots \mathbf{W}^{(\ell)} 
  \end{bmatrix}, \mathbf{b}^{(\ell)}_{split} = \begin{bmatrix} \mathbf{b}^{(\ell)}\dots\mathbf{b}^{(\ell)}\end{bmatrix},
\end{equation}
\begin{equation}
  \label{equ:outl}
  \mathbf{W}^{(L)}_{split} = \frac{1}{n}\begin{bmatrix} \mathbf{W}^{(L)} \dots \mathbf{W}^{(L)} \end{bmatrix}, \mathbf{b}^{(L)}_{split} = \frac{1}{n}\mathbf{b}^{(L)}.
\end{equation}
where the size of the vector and the number of columns in the matrix are equal to $n$.
\section{Numerical Experiments}
We will first evaluate the ability of the frequency upscaling to improve the efficiency of the training for higher-frequency wavefield representation. We then evaluate the ability of neuron splitting to help us get higher-accuracy for high-frequency wavefields. The tests are based on a simple layered model extracted from the Marmousi model (Figure~\ref{fig:2hz}(a)) covering an area of 2.5$\times$2.5 km$^2$. 
\subsection{PINN using frequency upscaling and neuron splitting}
\begin{figure}[!htb]
	\subfloat[]{\label{fig2a}}{\noindent\includegraphics[width=0.25\textwidth]{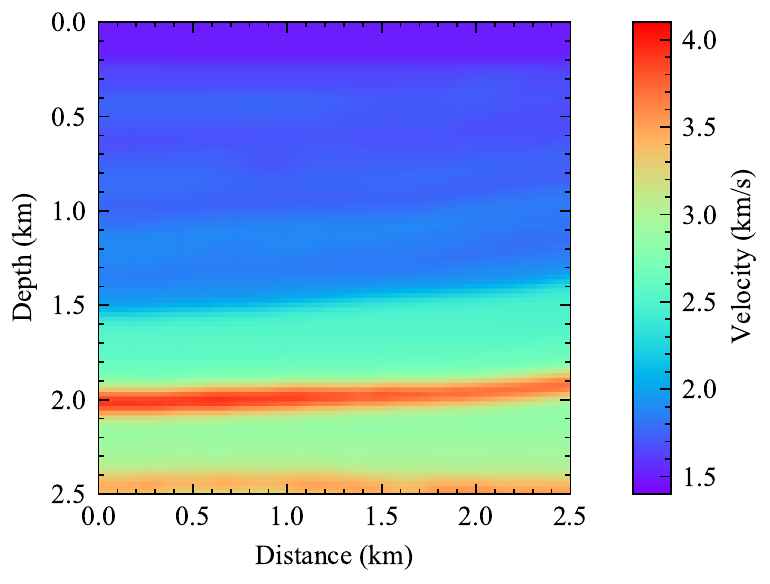}}
	\subfloat[]{\label{fig2b}}{\noindent\includegraphics[width=0.24\textwidth]{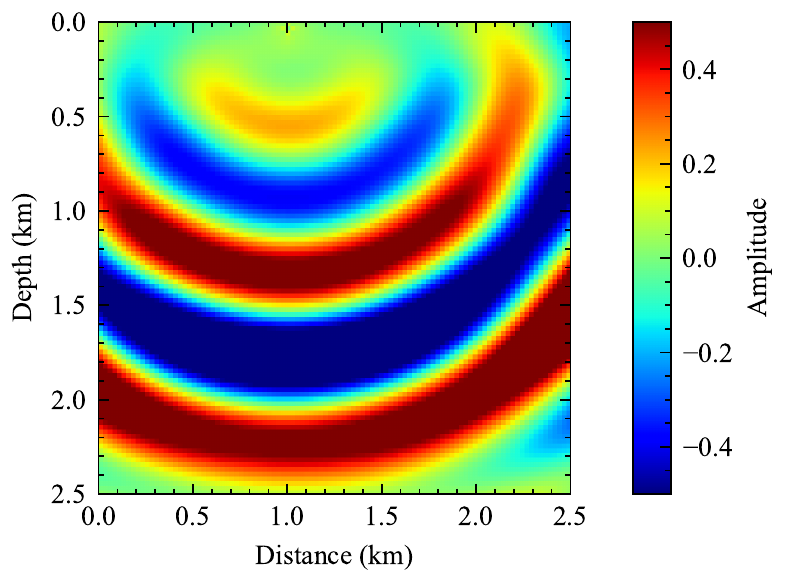}}
	\subfloat[]{\label{fig2c}}{\noindent\includegraphics[width=0.24\textwidth]{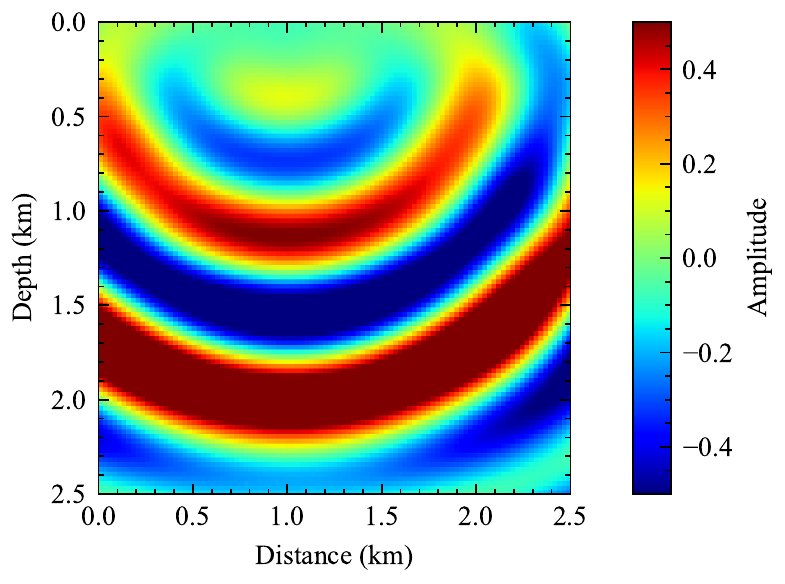}}
	\subfloat[]{\label{fig2d}}{\noindent\includegraphics[width=0.24\textwidth]{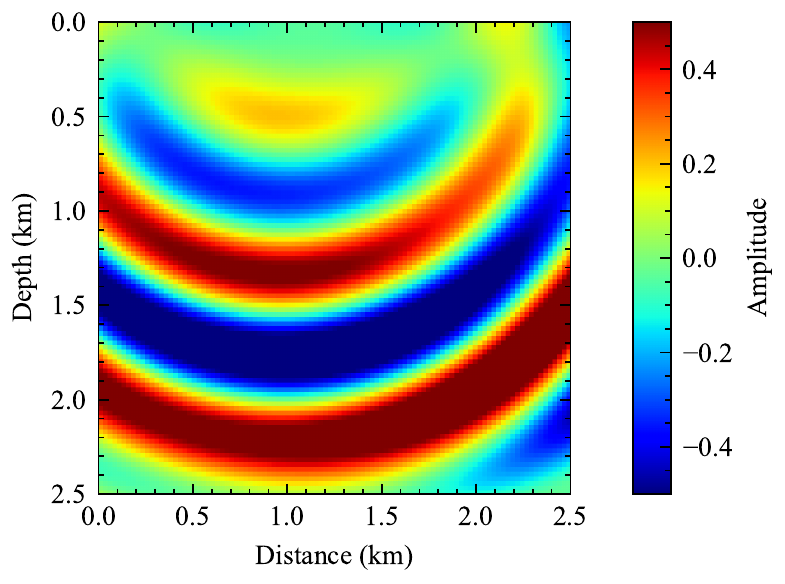}}
	\subfloat[]{\label{fig2e}}{\noindent\includegraphics[width=0.24\textwidth]{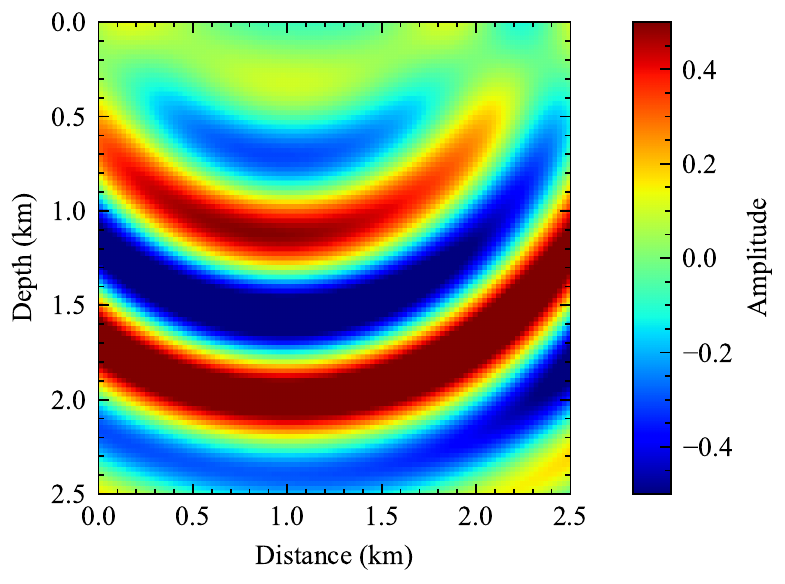}}
  \caption{The true velocity extracted from the Marmousi model (a), and the real and imaginary parts of the scattered wavefield for a source located at 1.0 km near the surface via a conventional numerical method (b, c) (considered ground truth) and PINN (d, e), respectively.}
	\label{fig:2hz}
\end{figure}

We first generated 10000 random training samples of $(x,z,x_s)$ covering the space coordinates of the wavefield, and the source location on the surface, with their corresponding $\delta\mathbf{m}$ for the squared slowness perturbation, and $\mathbf{m}_0$ for background squared slowness, needed for the loss function. The depth of sources $z_s$ is set to 0.025 km. The background wavefield is calculated analytically for a background velocity of 1.5 km/s. We train a small 2-layer fully-connected NN with $\{4,4\}$ neurons (4 neurons per layer) to predict the wavefield corresponding to a frequency of 2 Hz. The training is carried out over 50000 epochs using an Adam optimizer. The initial learning rate is 0.001 and decreases every 5000 epochs. To evaluate the results, we solve the Helmholtz equation numerically for a frequency of 2 Hz using the velocity model in Figure \ref{fig:2hz}(a). Using our trained NN, we then predict the solution on the same regular grid used for the numerical solution, which is discretized in the 2.5 $\times$ 2.5 km$^2$ area using 100 samples in both the $x$ and $z$ directions. The predictions are shown in Figure~\ref{fig:2hz}. The prediction (which is instant) of both the real and imaginary parts are consistent with the numerical solution. As a result, the 2 Hz wavefield for any source near the surface (Green's function) is stored as a function represented by 94 parameters. In other words, this 3D wavefield, which often requires at least 100$\times$100$\times$50 grid points to sample, is represented by less than 100 NN parameters.

\begin{figure}[!tb]
  \centering
  \subfloat[]{\label{fig3a}}{\noindent\includegraphics[width=0.24\textwidth]{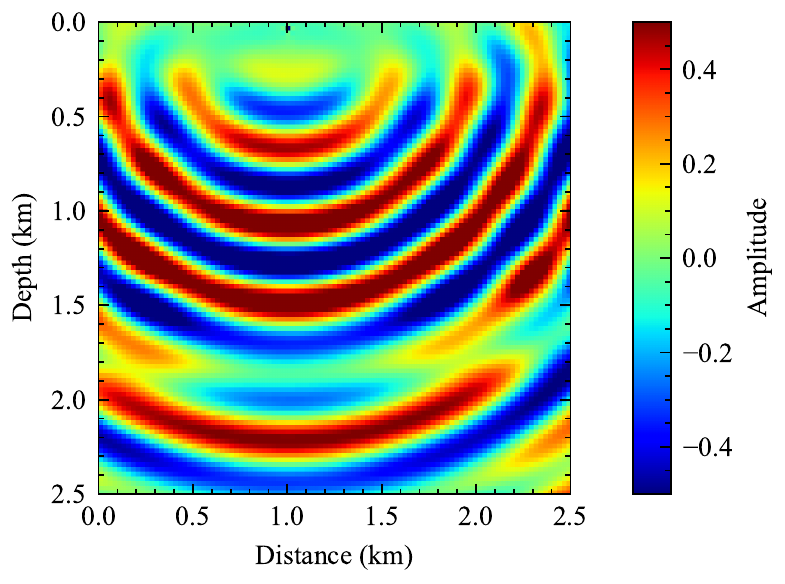}}
  \subfloat[]{\label{fig3b}}{\noindent\includegraphics[width=0.24\textwidth]{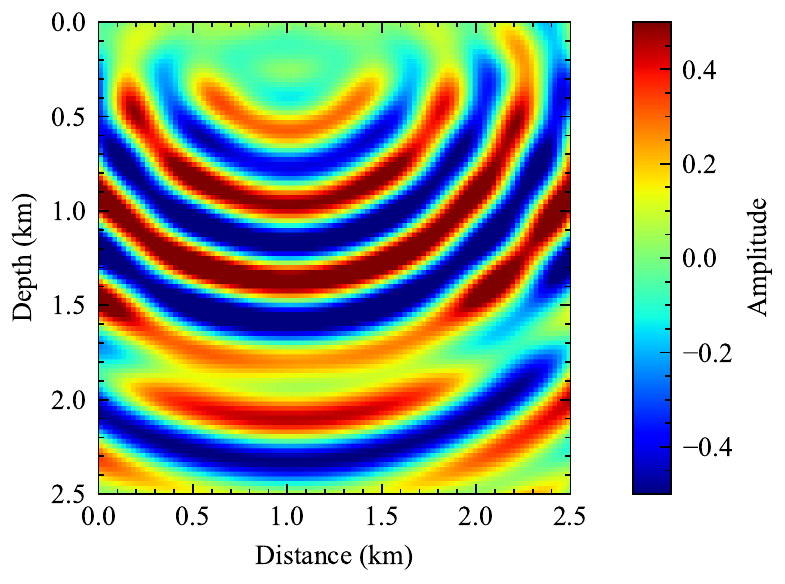}}
	\subfloat[]{\label{fig3c}}{\noindent\includegraphics[width=0.24\textwidth]{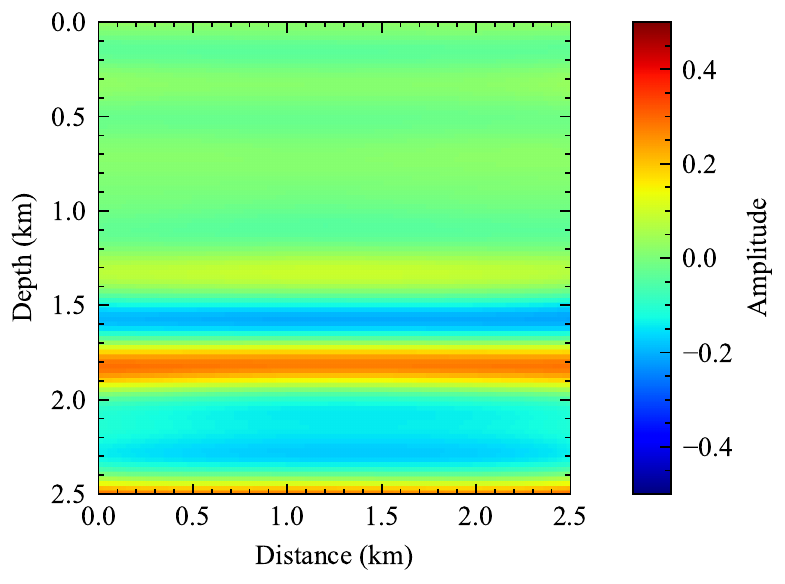}}
	\subfloat[]{\label{fig3d}}{\noindent\includegraphics[width=0.24\textwidth]{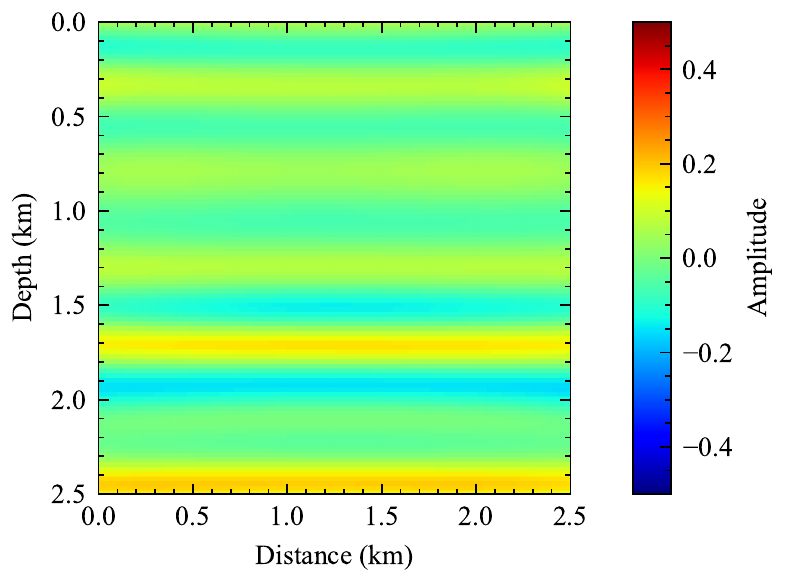}}
	\subfloat[]{\label{fig3e}}{\noindent\includegraphics[width=0.24\textwidth]{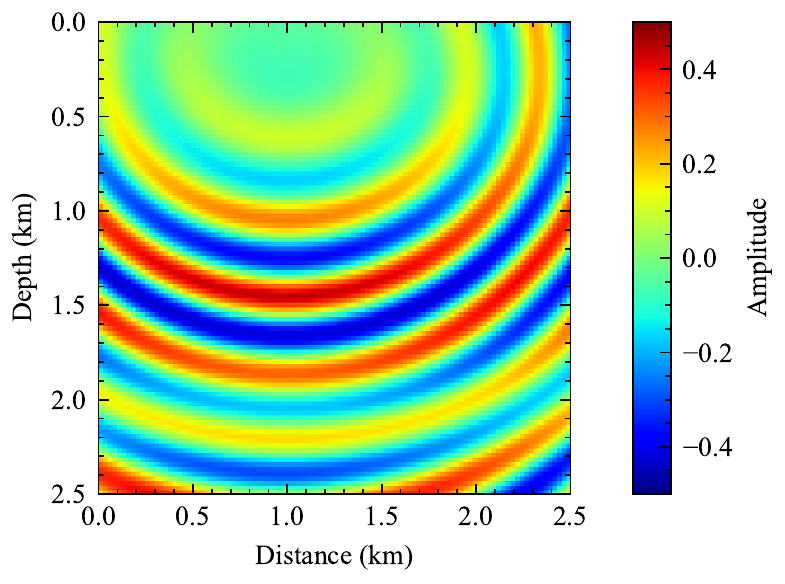}}
	\subfloat[]{\label{fig3f}}{\noindent\includegraphics[width=0.24\textwidth]{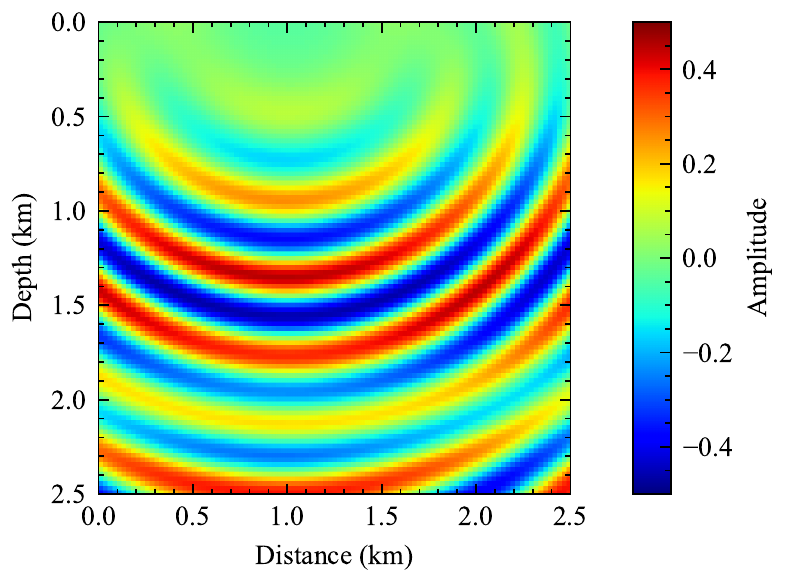}}
  \caption{The real (a, c, e) and imaginary (b, d, f) parts of the 4Hz scattered wavefield via a numerical method (a, b) and PINN using random initialization (c, d) and using the trained model for 2Hz (e, f).}
  \label{fig:4hz}
\end{figure}

\begin{figure}[!tb]
  \centering
  \includegraphics[width=0.48\textwidth]{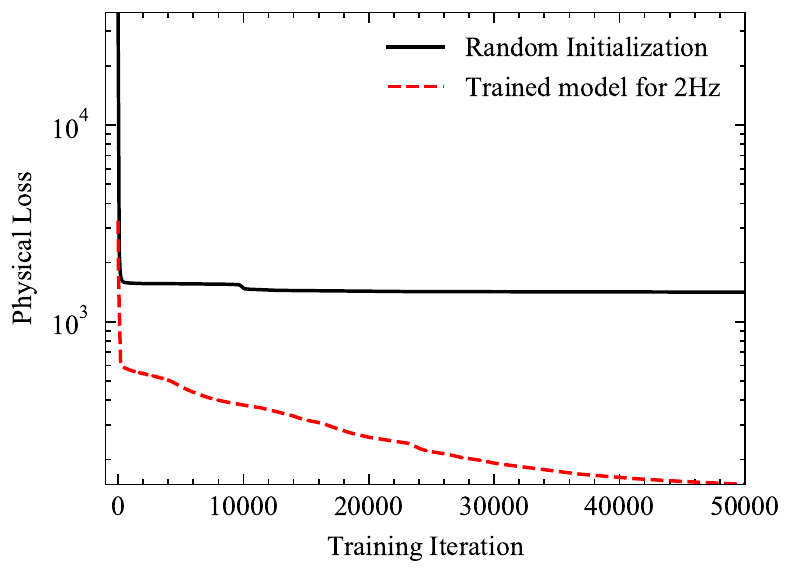}
  \caption{The comparison of the loss curves for training the model using random initialization and the model using trained model for 2 Hz.}
  \label{fig:4hz-loss}
\end{figure}

Then we train the same NN of $\{4,4\}$ neurons to predict a 4-Hz wavefield. The training parameters are the same as above but now we use 40000 randomly generated training samples. We train two independent NNs, with one of them using random initialization while the other using the pre-trained model on the 2-Hz wavefield as initial weights for the NN model. The predictions of these two models after 50000 epochs of training are shown in Figure~\ref{fig:4hz}. The prediction result of training with random initialization is erroneous, while initialization with the pre-trained model provides much better results, which includes more information of the 4 Hz wavefield. It demonstrates that the information gained from the previously trained model for 2-Hz helped accelerate the convergence of the 4Hz training. To illustrate this feature further, we share the loss functions for the two training tasks, as shown in Figure~\ref{fig:4hz-loss}. We can see that the PINN, even with positional encoding, needs a lot of epochs to start to converge when we begin from randomly initialized parameters. Moreover, the loss decreases slowly, which shows that a larger network may be needed in this case. In contrast, with the information gained from the previous training, the NN can better converge to the proper parameters. In other words, considering the same velocity model, NN information can transfer from low to high frequency. We still observe that even with this same small network, our proposed method was able to add additional information to the wavefield. This feature can help us utilize small more efficient networks in learning wavefield functions.

\begin{figure}[!tb]
  \centering
	\subfloat[]{\label{fig4a}}{\noindent\includegraphics[width=0.24\textwidth]{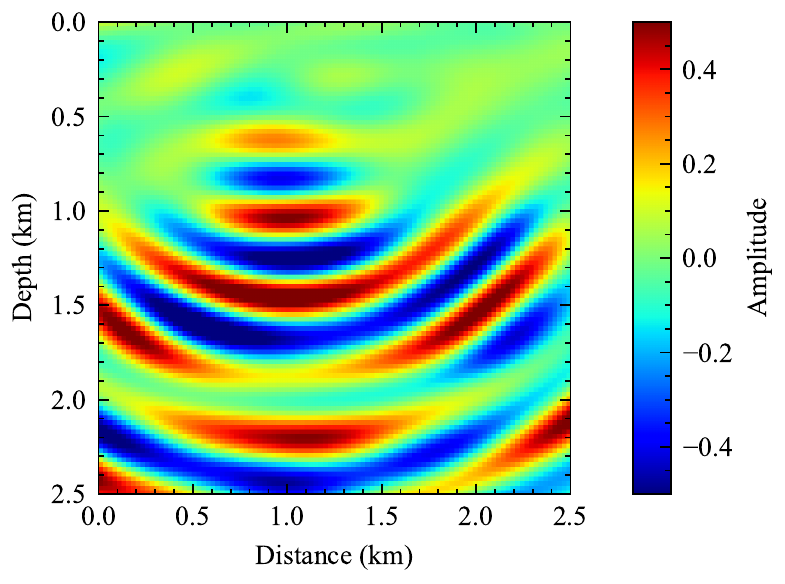}}
	\subfloat[]{\label{fig4b}}{\noindent\includegraphics[width=0.24\textwidth]{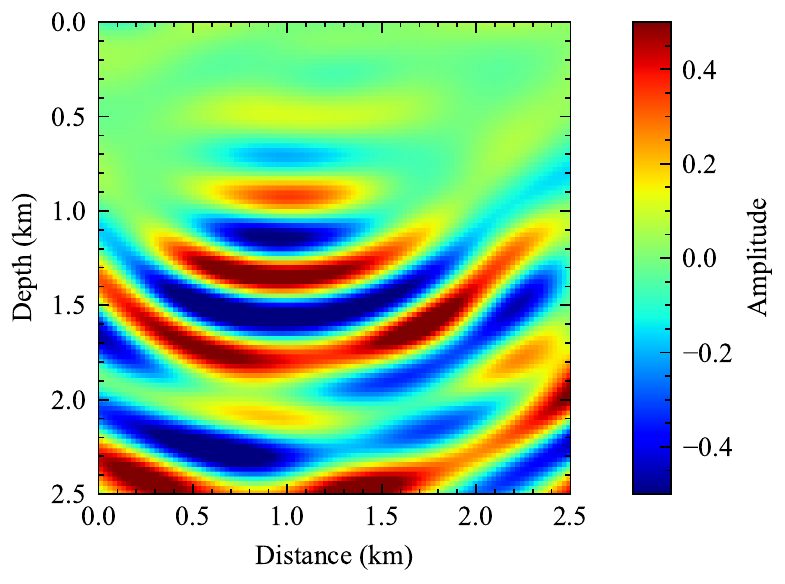}}
	\subfloat[]{\label{fig4c}}{\noindent\includegraphics[width=0.24\textwidth]{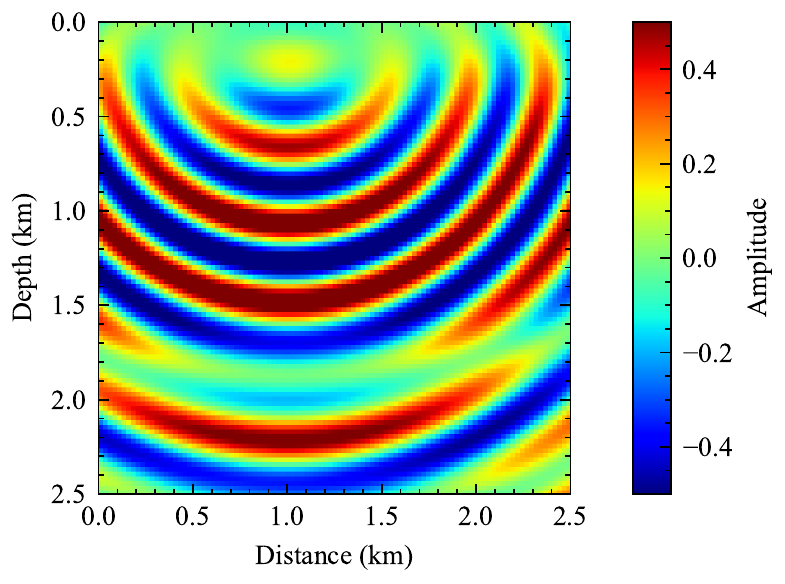}}
	\subfloat[]{\label{fig4d}}{\noindent\includegraphics[width=0.24\textwidth]{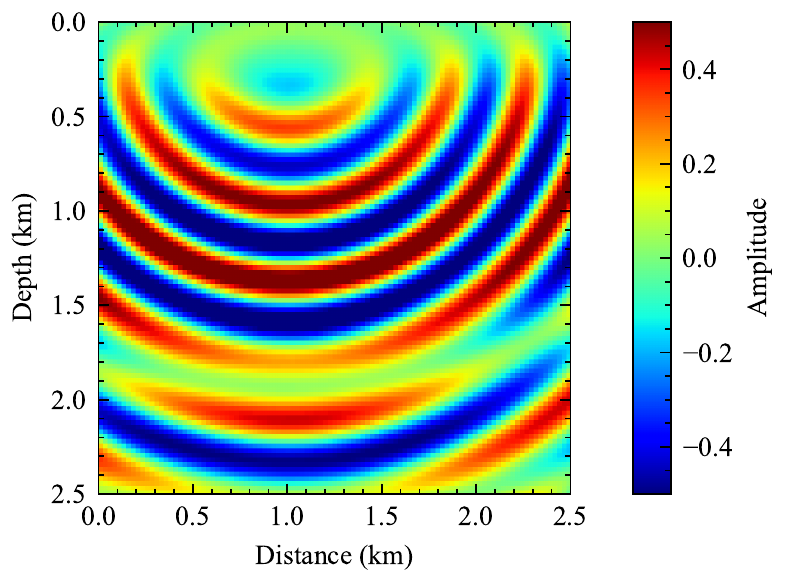}}
  \caption{The real (a, c) and imaginary (b, d) parts of the 4Hz scattered wavefield via a numerical method (a), PINN using random initialization (b) and using the trained model for 2Hz by neuron splitting (c).}
  \label{fig:4hz-ns}
\end{figure}

\begin{figure}
  \centering
  \includegraphics[width=0.48\textwidth]{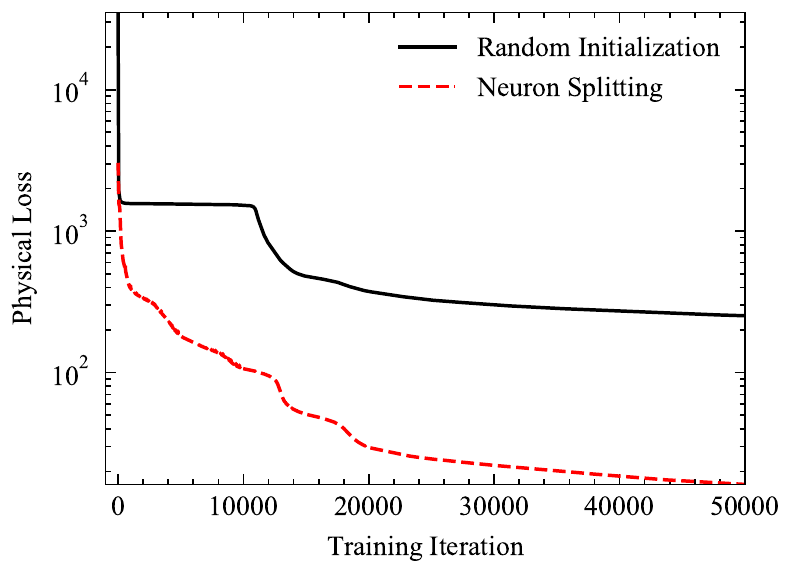}
  \caption{The comparison of the loss curves for training the model using random initialization and the model using trained model for 2 Hz by neuron splitting.}
  \label{fig:4hz-ns-loss}
\end{figure}

However, as shown in Figure~\ref{fig:4hz}(c), the predicted wavefield is highly smoothed compared with the numerical solution and some important features missing. The width of NN is too narrow to represent the higher-frequency wavefield, and as a result, it admitted a smooth version of it. Thus, we use neuron splitting to increase the NN size 4 times (we will discuss this choice of splitting in the Discussion section) compared to the trained model for 2Hz and use it for the additional training on the 4Hz wavefield. The prediction is shown in Figure~\ref{fig:4hz-ns}, which is much better than the prediction in Figure~\ref{fig:4hz}(c). We also compare the result with that trained with random initialization on the larger network and show their loss function (Figure~\ref{fig:4hz-ns-loss}). They demonstrate that PINN with frequency upscaling and neuron splitting can efficiently represent a new higher-frequency wavefield with higher accuracy.

\subsection{Accurate high-frequency wavefield}
Since we have observed the superiority of our proposed method in upscaling from 2 to 4 Hz, we next consider higher-frequency wavefields, which are more realistic and common in seismic applications. In this section and in following with FWI traditions \cite{Sirgue2004}, we upscale to three frequencies, 8 Hz, 16 Hz, 32 Hz in succession as examples, to test the performance of our proposed method in dealing with higher frequencies. We again increase the model size 4 times as the frequency is doubled and also increase the number of training samples by 4 times. The training strategies have been modified slightly here. We keep the batch size the same and decrease the epochs to keep the number of total iterations the same, which is more efficient. The source location of the evaluated wavefields is 1.0 km near the surface and the numerical solutions are considered as ground truth.

\begin{figure}[!tb]
  \centering
	\subfloat[]{\label{fig5a}}{\noindent\includegraphics[width=0.24\textwidth]{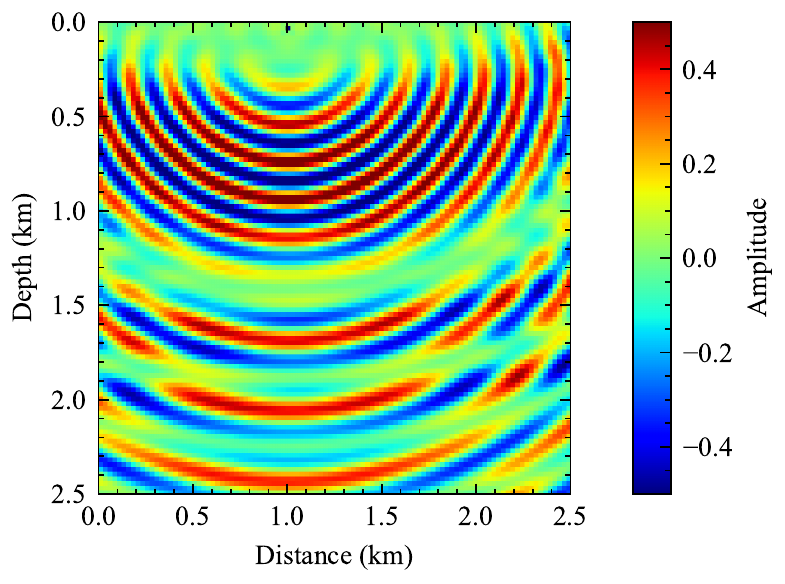}}
	\subfloat[]{\label{fig5b}}{\noindent\includegraphics[width=0.24\textwidth]{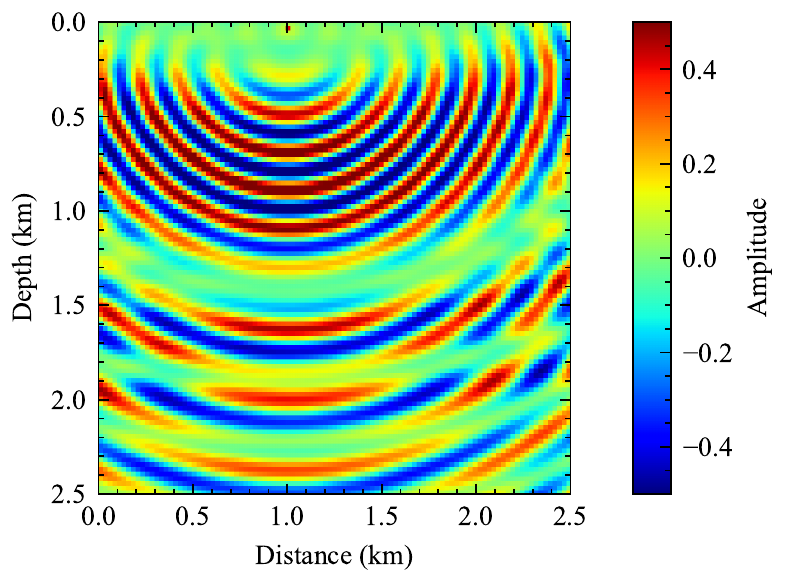}}
	\subfloat[]{\label{fig5c}}{\noindent\includegraphics[width=0.24\textwidth]{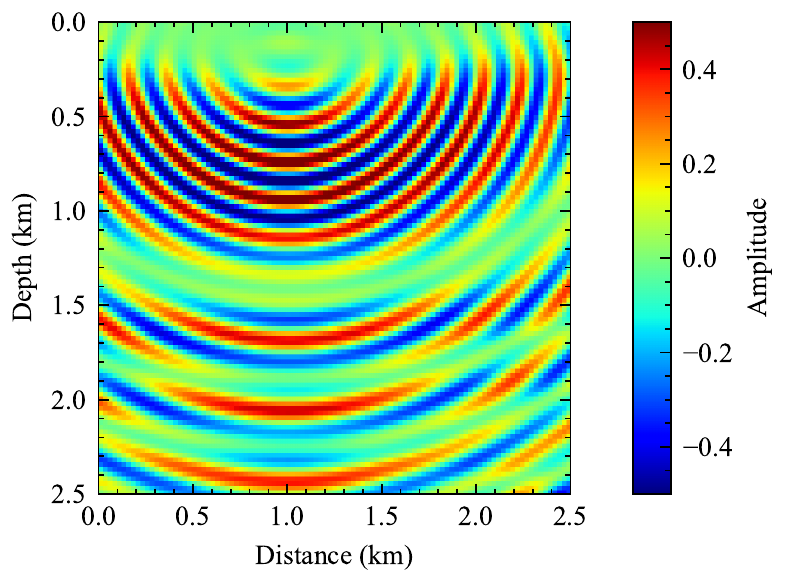}}
	\subfloat[]{\label{fig5d}}{\noindent\includegraphics[width=0.24\textwidth]{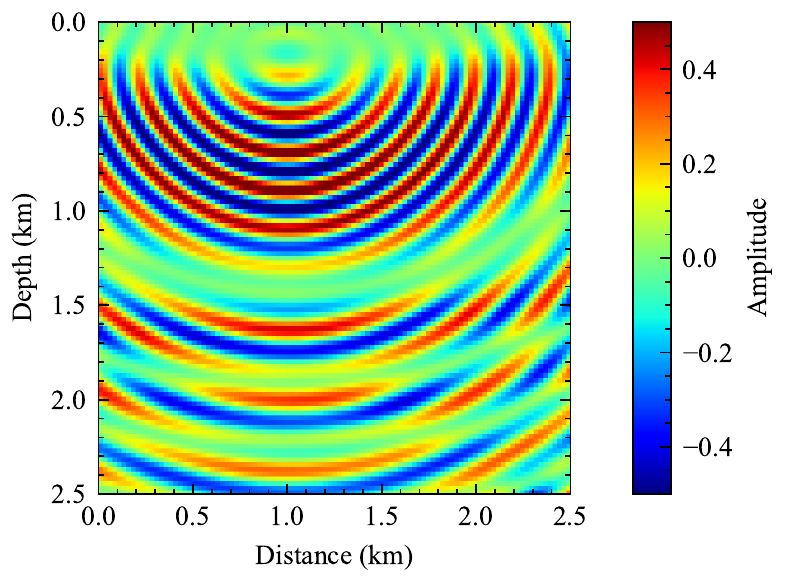}}
	\subfloat[]{\label{fig5e}}{\noindent\includegraphics[width=0.24\textwidth]{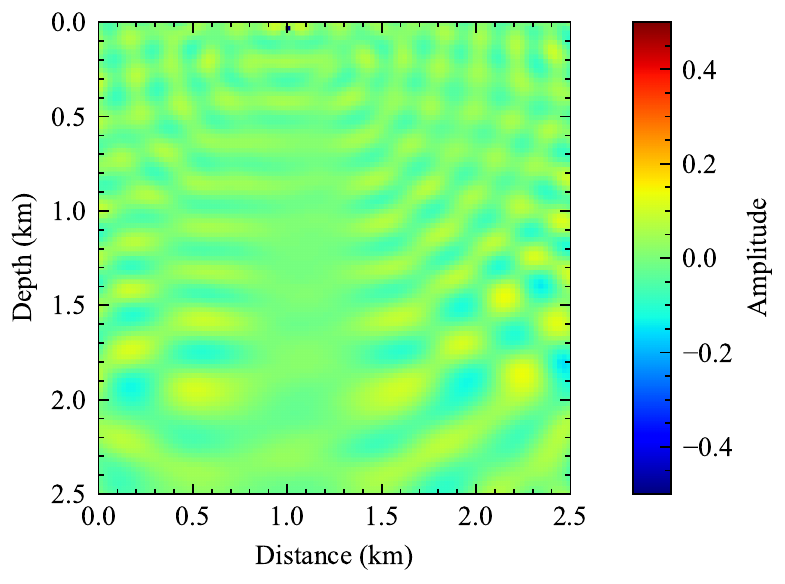}}
	\subfloat[]{\label{fig5f}}{\noindent\includegraphics[width=0.24\textwidth]{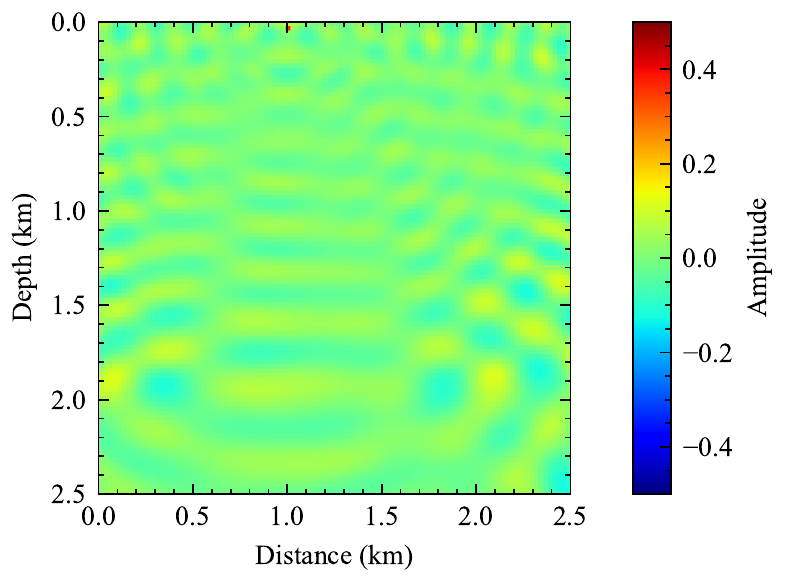}}
  \caption{The real (a, c, e) and imaginary (b, d, f) parts of the 8Hz scattered wavefield via a numerical method (a, b), PINN using the trained model for 4Hz by neuron splitting (c, d), and their corresponding differences (e, f).}
  \label{fig:8hz-ns}
\end{figure}

We train the 8 Hz wavefield based on 4 Hz model. Specifically, the network is initialized using the 4 Hz NN model and after we split each neuron to 4 times offsprings, resulting in a network with still two hidden layers, but 64 neurons in each layer. Figure~\ref{fig:8hz-ns} shows the predicted wavefields at 8 Hz. The PINNup solution recovers the almost every detail of 8 Hz wavefield for both the real and imaginary parts. The differences in the real and imaginary parts of the wavefields between the numerical solutions and the PINNup solutions are small, which proves that PINNup to 8 Hz worked fine. 

\begin{figure}[!htb]
  \centering
	\subfloat[]{\label{fig6a}}{\noindent\includegraphics[width=0.24\textwidth]{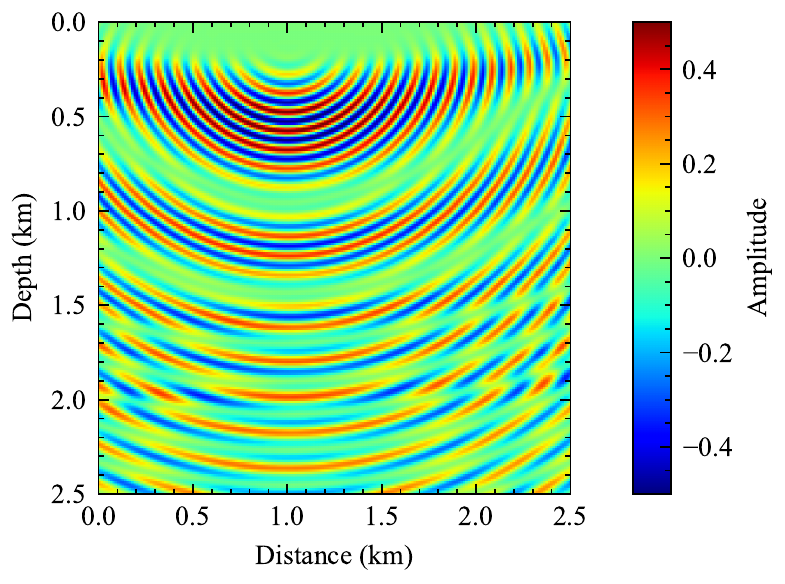}}
	\subfloat[]{\label{fig6b}}{\noindent\includegraphics[width=0.24\textwidth]{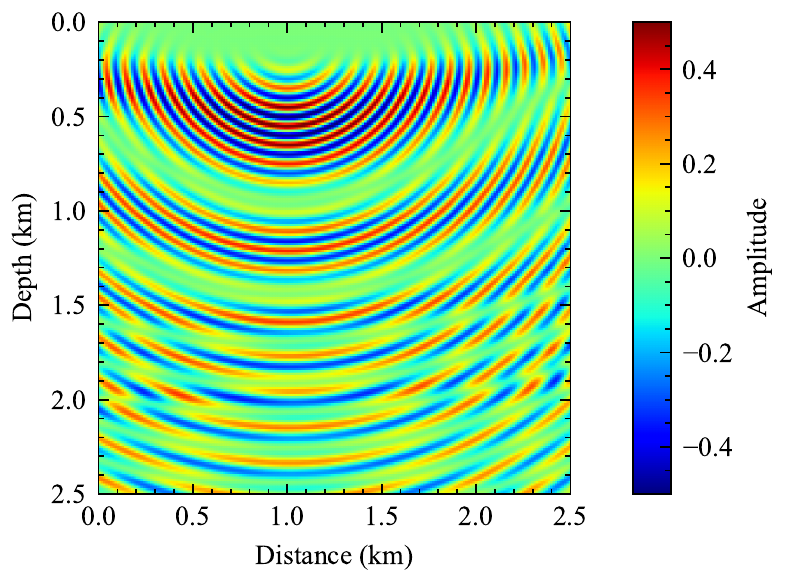}}
	\subfloat[]{\label{fig6c}}{\noindent\includegraphics[width=0.24\textwidth]{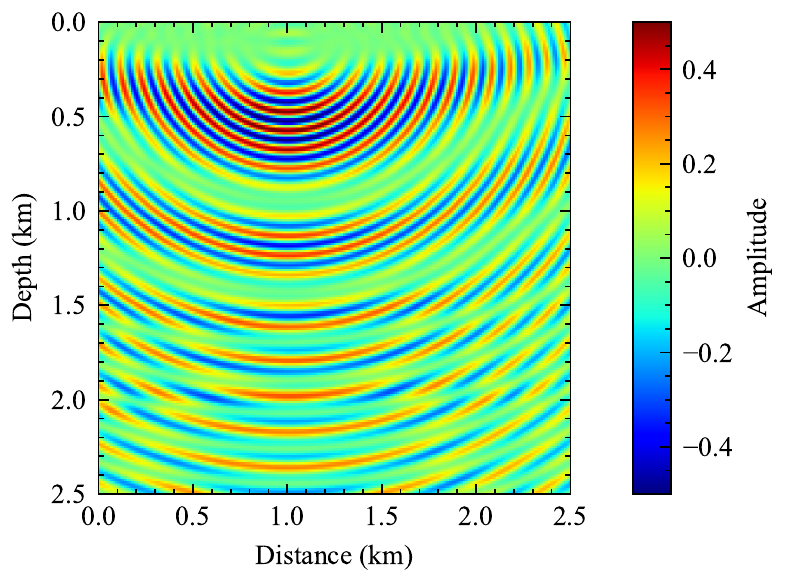}}
	\subfloat[]{\label{fig6d}}{\noindent\includegraphics[width=0.24\textwidth]{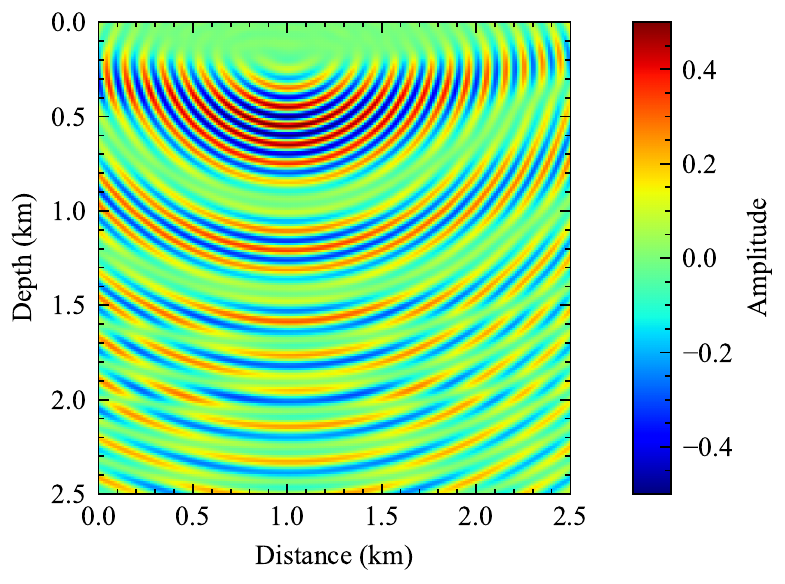}}
	\subfloat[]{\label{fig6e}}{\noindent\includegraphics[width=0.24\textwidth]{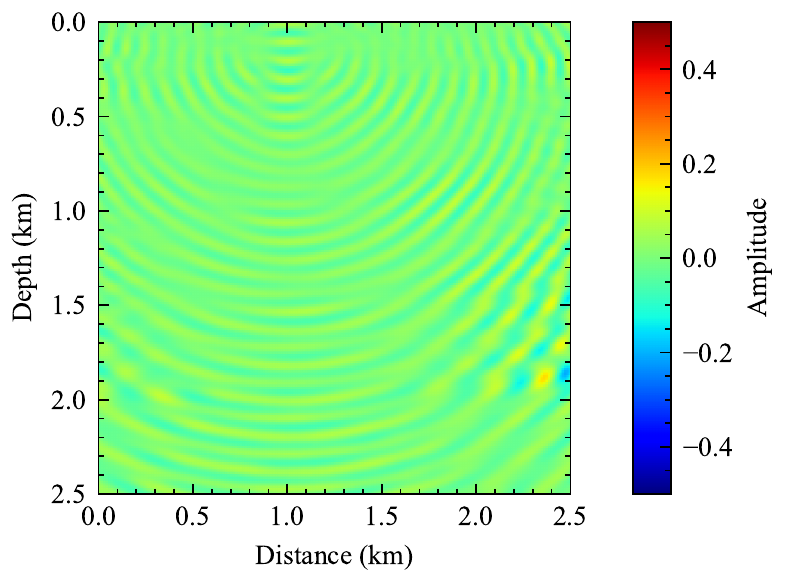}}
	\subfloat[]{\label{fig6f}}{\noindent\includegraphics[width=0.24\textwidth]{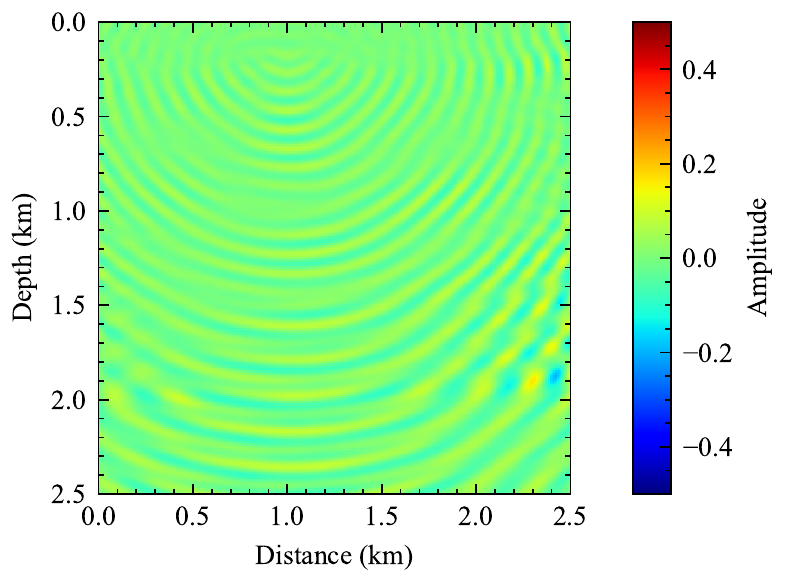}}
  \caption{The real (a, c, e) and imaginary (b, d, f) parts of the 16Hz scattered wavefield via a numerical method (a, b), PINN using the trained model for 8Hz by neuron splitting (c, d), and their corresponding differences (e, f).}
  \label{fig:16hz-ns}
\end{figure}

Then we train the 16 Hz wavefield using 8 Hz pre-trained model in the above experiment. After splitting, the neural network has two-hidden-layers with 256 neurons in each layer. For the numerical solution at 16 Hz, we use a finer grid of 200 samples in both the $x$ and $z$ directions to avoid numerical dispersion. As shown in Figure~\ref{fig:16hz-ns}, the solution provided by PINNup is close to the numerical one, as it captures the complex features of the 16 Hz wavefield. The difference between the numerical solution and PINNup solution is also small. 

\begin{figure}[!tb]
  \centering
	\subfloat[]{\label{fig7a}}{\noindent\includegraphics[width=0.24\textwidth]{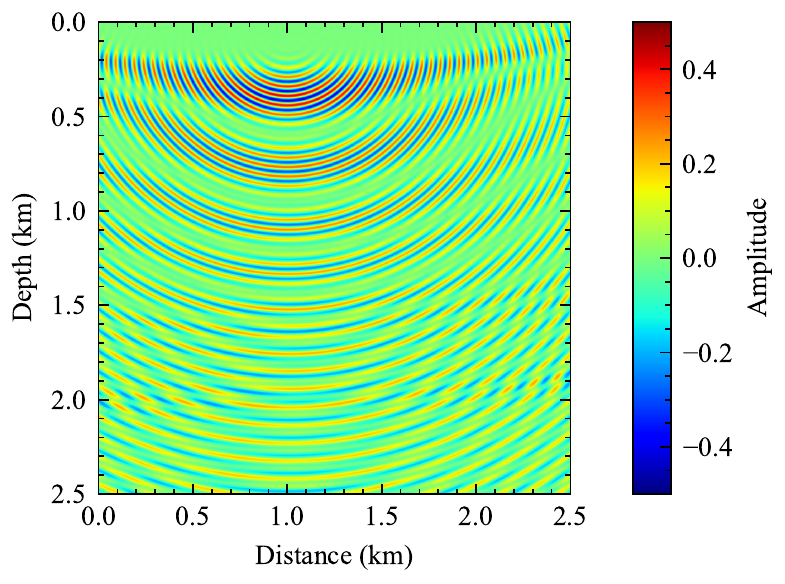}}
	\subfloat[]{\label{fig7b}}{\noindent\includegraphics[width=0.24\textwidth]{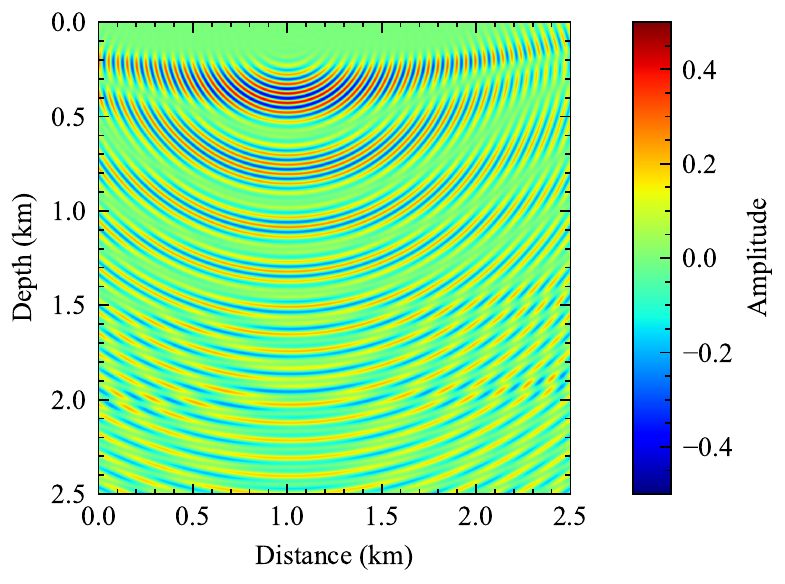}}
	\subfloat[]{\label{fig7c}}{\noindent\includegraphics[width=0.24\textwidth]{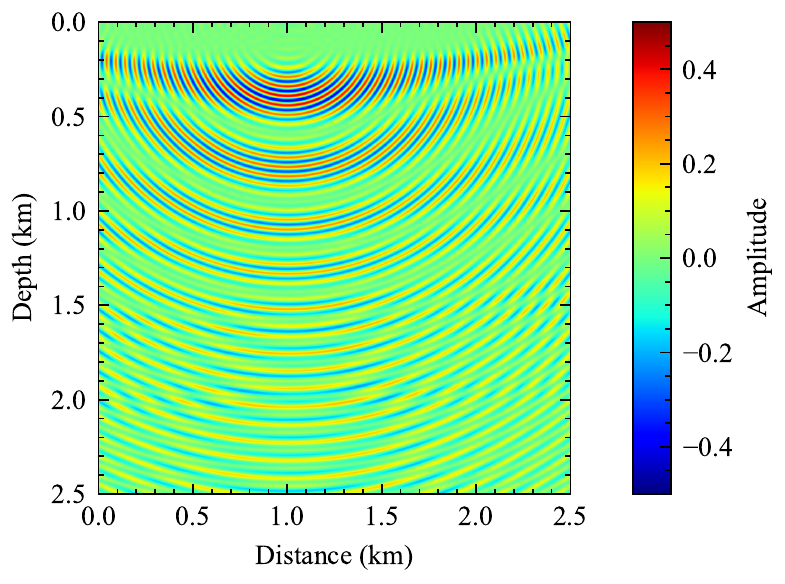}}
	\subfloat[]{\label{fig7d}}{\noindent\includegraphics[width=0.24\textwidth]{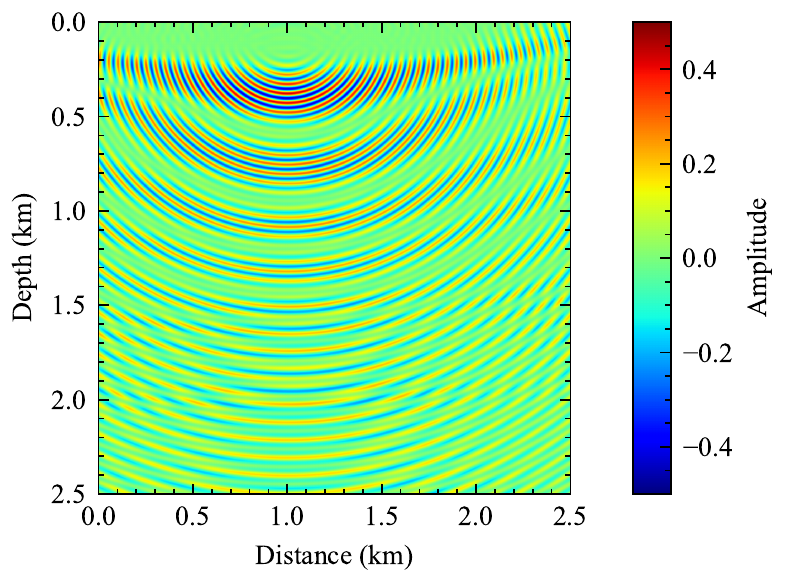}}
	\subfloat[]{\label{fig7e}}{\noindent\includegraphics[width=0.24\textwidth]{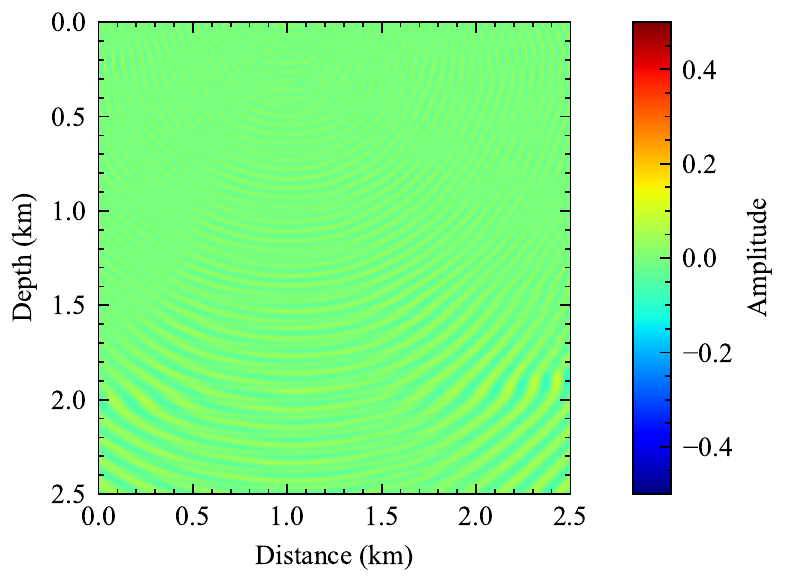}}
	\subfloat[]{\label{fig7f}}{\noindent\includegraphics[width=0.24\textwidth]{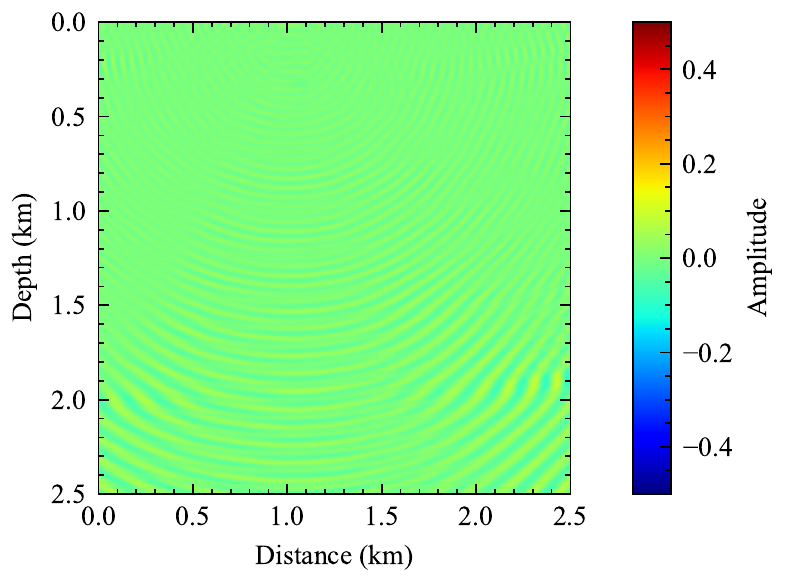}}
  \caption{The real (a, c, e) and imaginary (b, d, f) parts of the 32Hz scattered wavefield via a numerical method (a, b), PINN using the trained model for 16Hz by neuron splitting (c, d), and their corresponding differences (e, f).}
  \label{fig:32hz-ns}
\end{figure}

Finally, we train the 32 Hz wavefield starting with the previous model for 16 Hz wavefield. The network size after repeating the same splitting strategy described above becomes $\{1024,1024\}$. Again, to avoid numerical dispersion in numerical solutions, which is used as reference, we use a finer grid of 800 samples in each of the $x$ and $z$ directions. The results of PINNup for the real and imaginary parts of the 32 Hz wavefield are shown in Figures~\ref{fig:32hz-ns}(c) and (d). Its phase and amplitude features are very close to the numerical solutions. We display the difference plots and find that the representation of 32 Hz wavefield has good accuracy.

Even though the representations of wavefield via NN at every frequency may include some errors, those errors, as we saw, are generally small, and we managed to capture the most important features of the wavefield. To further highlight the importance of our approach for PINN, we refer the reader to \cite{Sitzmann2020} in which they use a sine activation function to predict a 3.2 Hz wavefield corresponding to Helmholtz equation, and to do so they needed a network of 5 layers with 256 neurons per layer, and that prediction was for a single source.

From the above experiments, we observe that with our proposed method, we can easily predict high-accuracy multi-frequency wavefield solutions at any location in the domain of interest corresponding to any source location (because the input of our network includes the coordinates of source location and arbitrary space coordinates). In other words, no interpolation is needed. Such a continuous wavefield representation is stored in a much more compressed form even for the 32 Hz representation.

\section{Discussions}
The purpose of this work is to demonstrate the potential to upscale from low to higher frequency in training neural network wavefield functions. The frequency upscaling improved the convergence of PINN, as it leverages key information from the lower-frequency in the training, which is an optimization problem. A similar feature is noticed in FWI as we build velocity models. This is consistent with the low-frequency bias property of NN. In other words, the NN tends to focus on the low-frequency component of a function first and then adds higher-frequency components gradually. Thus, the pre-trained model from the low-frequency wavefield can be a good initial model for higher frequency training. As a result, we simply need some additional training to transfer the knowledge of predicting low-frequency wavefields to higher-frequency ones. However, to fully benefit from this feature, we also utilize here the concept of neuron splitting to increase the size of our NN model to reflect the complexity of the predicted wavefields at higher frequency. This allows us to predict higher accuracy wavefields at a reduced cost. The splitting parameter $n$ depends on the frequency upscaling. We found empirically, as the examples show, that if we initialize our training with the NN model parameters used to predict a wavefield at a frequency half that of interest, we need to split the neurons by 4. This number is directly proportional to the effective increase of doubling the frequency on the grid requirement for each of the two axis in 2D (2$\times$2). So in 3D media, we assume that the required splitting for a prediction at a frequency that is double that of the previously trained network would be 8 (2$\times$2$\times$2). The same ratios hold for the number of training samples. In other words, this reasoning comes from the fact that for the same velocity model in 2D, a 2 Hz wavefield in a 4 km$^2$ area is equivalent in complexity to a 4 Hz wavefield in a 1 km$^2$ area. We noticed that the third dimension given by the source has little baring on this empirical relation, because if the velocity is laterally invariant, the wavefields from the various sources are similar. Thus, more complex lateral variations might require us to split the neurons even more. One interesting component of this upscaling is that, unlike the numerical solutions where a doubling of the frequency implies sampling that increase the cost by 10 times in 2D, the splitting and the increase in the training samples for higher frequency increases the cost by around 6 times. 

Moreover, the neuron splitting we incorporated is one approach from a family of approaches. We can also use optimized neuron splitting to make the NN architecture for higher-frequency wavefields more adaptable to the prediction needs \cite{liu2019splitting}. In this case, the splitting is not fixed for every neuron, as it depends on the optimization. However, the down side to this approach is the additional cost of this adaptive splitting. In most cases, we believe the fixed neuron splitting guided by the frequency upscaling is good enough.

\section{Conclusions}
We proposed an efficient PINN resulting in higher-accuracy in predicting wavefields at higher frequencies courtesy of frequency upscaling and neuron splitting. We train a small NN with a small number of training samples to predict low frequency wavefields. The trained NN model is used as an initial model for training at higher frequencies using a larger neural network architecture by utilizing neuron splitting. The convergence and accuracy of this approach in predicting wavefields at higher frequencies exceeds those obtained through random initialization of the NN model. Applications on a simple layered model demonstrated such features. Even with a shallow network, we can still leverage the proposed method to get high-frequency wavefields with good accuracy, which is very hard for conventional PINN. These features are very favorable for future applications in waveform inversion.

\section*{Acknowledgment}
We thank KAUST for its support and the SWAG group for the collaborative environment. This work utilized the resources of the Supercomputing Laboratory at King Abdullah University of Science and Technology (KAUST) in Thuwal, Saudi Arabia. 
\ifCLASSOPTIONcaptionsoff
  \newpage
\fi

\bibliographystyle{IEEEtran}
\bibliography{Pupf.bib}
%
\end{document}